\documentclass{article}

\usepackage{xcolor,soul}
\sethlcolor{yellow}
\usepackage{arxiv}
\usepackage{hyperref}       
\usepackage{url}            
\usepackage{amsfonts}       
\usepackage{nicefrac}       
\usepackage{authblk}
\usepackage{graphicx}
\usepackage{multirow}
\usepackage{subfig}
\usepackage{mathtools}
\usepackage{algorithm,algpseudocode}
\title{A deep active learning system for species identification and counting in camera trap images}

\author[1,2]{\textbf{Mohammad Sadegh Norouzzadeh}}
\author[1]{\textbf{Dan Morris}}
\author[1,4]{\textbf{Sara Beery}}
\author[3]{\textbf{Neel Joshi}}
\author[3]{\textbf{Nebojsa Jojic}}
\author[2,5]{\textbf{Jeff Clune}}
\affil[1]{Microsoft AI for Earth, Redmond, WA}
\affil[2]{Computer Science department, University of Wyoming, Laramie, WY}
\affil[3]{Microsoft Research, Redmond, WA}
\affil[4]{Computer Science Department, California Institute of Technology, Pasadena, CA}
\affil[5]{Uber AI, San Francisco, CA}

\begin{document}
\maketitle

\begin{abstract}
Biodiversity conservation depends on accurate, up-to-date information about wildlife population distributions. Motion-activated cameras, also known as camera traps, are a critical tool for population surveys, as they are cheap and non-intrusive. However, extracting useful information from camera trap images is a cumbersome process: a typical camera trap survey may produce millions of images that require slow, expensive manual review. Consequently, critical information is often lost due to resource limitations, and critical conservation questions may be answered too slowly to support decision-making. Computer vision is poised to dramatically increase efficiency in image-based biodiversity surveys, and recent studies have successfully harnessed deep learning techniques for automatic information extraction from camera trap images. However, the accuracy of results depends on the amount, quality, and diversity of the data available to train models, and the literature has focused on projects with millions of relevant, labeled training images. Many camera trap projects do not have a large set of labeled images, and hence cannot benefit from existing machine learning techniques. Furthermore, even projects that do have labeled data from similar ecosystems have struggled to adopt deep learning methods because image classification models overfit to specific image backgrounds (i.e., camera locations). In this paper, we focus not on \textit{automating} the labeling of camera trap images, but on \textit{accelerating} this process. We combine the power of machine intelligence and human intelligence to build a scalable, fast, and accurate active learning system to minimize the manual work required to identify and count animals in camera trap images. Our proposed scheme can match the state of the art accuracy on a 3.2 million image dataset with as few as 14,100 manual labels, which means decreasing manual labeling effort by over 99.5\%.
\end{abstract}


\section{Introduction}
Wildlife population studies depend on tracking observations, i.e. occurrences of animals at recorded times and locations. This information facilitates the modeling of population sizes, distributions, and environmental interactions \cite{elith2010art,tikhonov2017using,southwood2009ecological}. Motion-activated cameras, or “camera traps”, provide a non-intrusive and comparatively cheap method to collect observational data, and have transformed wildlife ecology and conservation in recent decades \cite{o2010camera,burton2015wildlife}. Although camera trap networks can collect large volumes of images, turning raw images into actionable information is done manually, i.e. human annotators view and label each image \cite{swanson2015snapshot}. The burden of manual review is the main disadvantage of camera trap surveys and limits the use of camera traps for large-scale studies.  

Fortunately, recent advances in artificial intelligence have significantly accelerated information extraction. Loosely inspired by animal brains, deep neural networks \cite{lecun2015deep,goodfellow2016deep} have advanced the state of the art in tasks such as machine translation \cite{cho2014learning,sutskever2014sequence}, speech recognition \cite{bahdanau2016end,hinton2012deep}, and image classification \cite{he2016deep,simonyan2014very}. Deep convolutional neural networks are a class of deep neural networks designed specifically to process images \cite{goodfellow2016deep,krizhevsky2012imagenet}. 

Recent work has demonstrated that deep convolutional neural networks can achieve a high level of accuracy in extracting information from camera trap images---including species labels, count, and behavior---while being able to process hundreds of images in a matter of seconds \cite{norouzzadeh2018automatically,tabak2018machine}. The wide availability of deep learning for fast, automatic, accurate, and inexpensive extraction of such information could save substantial amounts of time and money for conservation biologists.

The accuracy of deep neural networks depends on the abundance of their training data \cite{goodfellow2016deep}; state-of-the-art networks typically require millions of labeled training images.  This volume of labeled data is not typically available for camera trap projects; therefore, most projects cannot yet effectively harness deep learning. Even in cases where an extensive training set is available, training labels are almost always in the form of image-level or sequence-level species labels, i.e. they do not contain information about where animals occur within each image. This results in a strong dependency of deep networks on image backgrounds \cite{miao2019insights,beery2018rec}, which limits the ability of deep learning models to produce accurate results even when applied to regions with species distributions that are similar to their training data, but with different backdrops due to different camera trap locations.

This paper aims to address these issues and to enable camera trap projects with few labeled images to take advantage of deep neural networks for fast, transferable, automatic information extraction. Using object detection models, transfer learning, and active learning, our results show that our suggested method can achieve the same level of accuracy as a recent study by Norouzzadeh et al. \cite{norouzzadeh2018automatically} that harnessed 3.2 million labeled training examples to produce 90.9\% accuracy (using ResNet-50 architecture) at species classification, but with a 99.5\% reduction in manually-annotated training data. We also expect our method to generalize better to new locations because we systematically filter out the background pixels.

\section{Background and related work}
\label{sec:headings}

\subsection{Deep learning}
The most common type of machine learning used for image classification is \textit{supervised learning}, where input examples are provided along with corresponding output examples (for example, camera trap images with species labels), and algorithms are trained to translate inputs to the appropriate outputs \cite{mohri2012foundations}. 

\textit{Deep learning} is a specific type of supervised learning, built around \textit{artificial neural networks} \cite{hagan1996neural,goodfellow2016deep}, a class of machine learning models inspired by the structure of biological nervous systems.  Each artificial neuron in a network takes in several inputs, computes a weighted sum of those inputs, passes the result through a non-linearity (e.g. a sigmoid), and transmits the result along as input to other neurons. Neurons are usually arranged in several layers; neurons of each layer receive input from the previous layer, process them, and pass their output to the next layer. A \textit{deep} neural network is a neural network with three or more layers \cite{goodfellow2016deep}. Typically, the free parameters of the model that are trained are the \emph{weights} (aka connections) between neurons, which determine the weight of each feature in the weighted sum.

In a \textit{fully-connected layer}, each neuron receives input from all the neurons in the previous layer. On the other hand, in \textit{convolutional layers}, each neuron is only connected to a small group of nearby neurons in the previous layer and the weights are trained to detect a useful pattern in that group of neurons \cite{hagan1996neural,goodfellow2016deep}. Additionally, convolutional neural networks inject the prior knowledge that translation invariance is helpful in computer vision (e.g. an eye in one location in an image remains an eye even if it appears somewhere else in the image). This is enforced by having a feature detector reused at many points throughout the image (known as \emph{weight tying} or \emph{weight sharing}. A neural network with one or more convolutional layers is called a \textit{convolutional neural network}, or CNN. CNNs have shown excellent performance on image-related problems \cite{lecun2015deep,goodfellow2016deep}. 

The weights of a neural network (aka its parameters) determine how it translates its inputs into outputs; \textit{training} a neural network means adjusting these parameters for every neuron so that the whole network produces the desired output for each input example. To tune these parameters, a measure of the discrepancy between the current output of the network and the desired output is computed; this measure of discrepancy is called the \textit{loss function}. There are numerous loss functions used in the literature that are appropriate for different problem classes. After calculating the loss function, an algorithm called Stochastic Gradient Descent (SGD) \cite{robbins1951stochastic,hecht1989theory} (or modern enhancements of it \cite{kingma2014adam,tieleman2012lecture}) calculates the contribution of each parameter to the loss value, then adjusts the parameters so that the loss value is minimized. The backpropagation algorithm is an iterative algorithm, i.e. it is applied many times during training, including multiple times for each image in the dataset. At every iteration of the backpropagation algorithm, the parameters take one step toward a minimum (i.e. the best solution in a local area of the search space of all possible weights: note that the term minima is used instead of maxima because we are minimizing the loss, or the error). 

The accuracy of deep learning compared to other machine learning methods makes it applicable to a variety of complex problems. In this paper, we focus on enhancing deep neural networks to extract information from camera trap images more efficiently. 

\subsection{Image classification}
In the computer vision literature, \textit{image classification} refers to assigning images into several pre-determined classes. More specifically, image classification algorithms typically assign a probability that an image belongs to each class. For example, species identification in camera trap images is an image classification problem in which the input is the camera trap image and the output is the probability of the presence of each species in the image \cite{norouzzadeh2018automatically,tabak2018machine}. Image classification models can be easily trained with image-level labels, but they suffer from several limitations:

\begin{enumerate}
\item Typically the most probable species is considered to be the label for the image; consequently, classification models cannot deal with images containing more than one species.
\item Applying them to non-classification problems like counting results in worse performance than classification \cite{norouzzadeh2018automatically}.
\item What the image classification models see during training are the images and their associated labels; they have not been told what \textit{parts} of the images they should focus on. Therefore, they not only learn about patterns representing animals, but will also learn some information about \textit{backgrounds} \cite{miao2019insights}. This fact limits their transferability to new locations. Therefore, when applied to new datasets, accuracy is typically lower than what was achieved on the training data. For example, Tabak et al. \cite{tabak2018machine} showed that their model trained on images from the United States was less accurate at identifying the same species in a Canadian dataset.
\end{enumerate}

\subsection{Object detection}
\begin{figure}[h]
	\setlength{\unitlength}{0.5in}
	\centering 
	\includegraphics[width=0.70\textwidth]{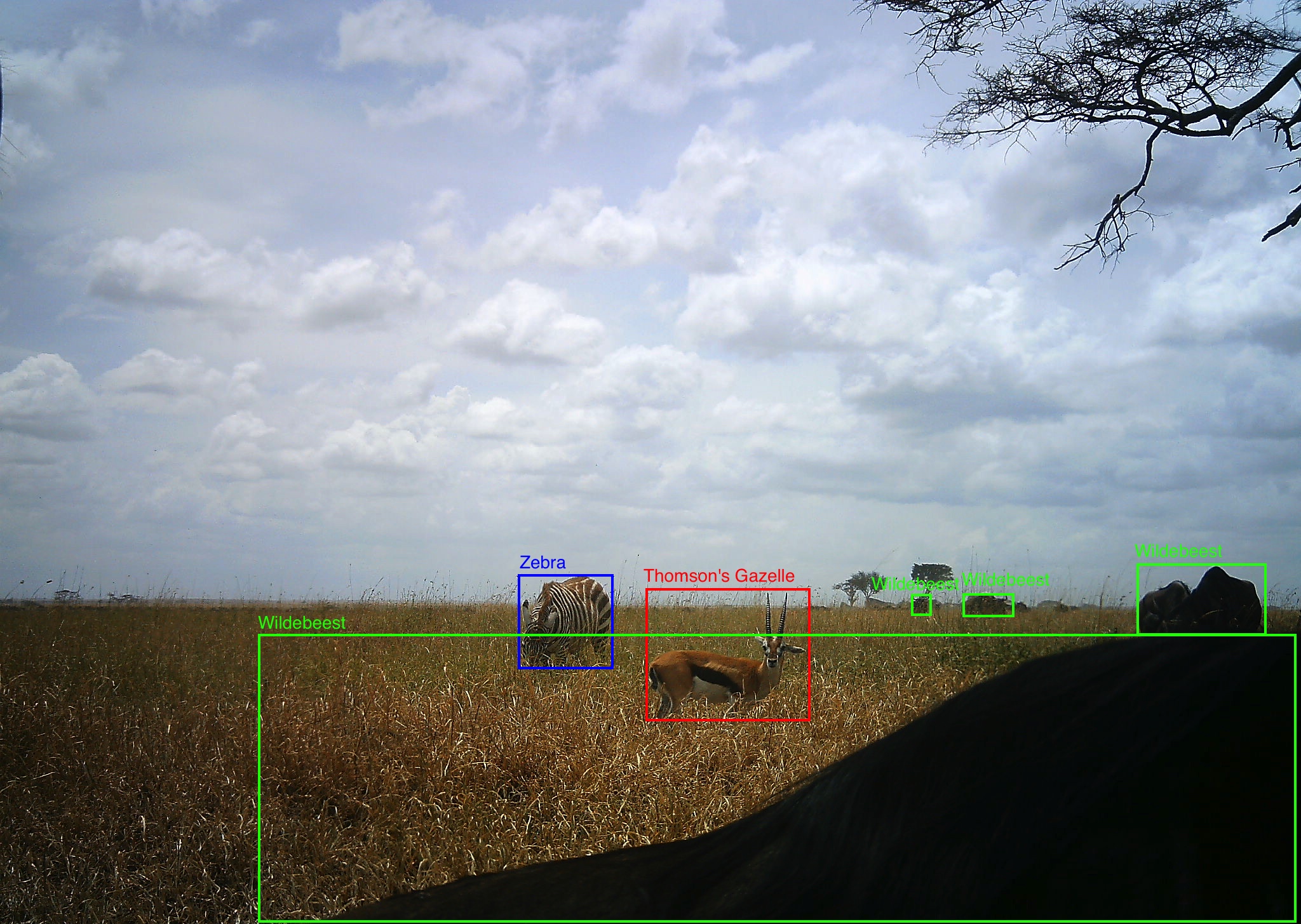}
	\caption{Object detection models are capable of detecting multiple occurrences of several object classes. 
	}
	\label{fig:example}
\end{figure}

\textit{Object detection} algorithms attempt to not only classify images, but to locate instances of predefined object classes within images.  Object detection models output coordinates of bounding boxes containing objects plus a probability that each box belongs to each class. Object detection models thus naturally handle images with objects from multiple classes. (Fig. \ref{fig:example}). A hypothesis of this paper is that object detection models may also be less sensitive to image backgrounds (because the model is told explicitly which regions of each image to focus on), and may thus generalize more effectively to new locations.

The ability of object detection models to handle images with multiple classes makes them appealing for camera trap problems, where multiple species may occur in the same images. However, training object detection models requires bounding box and class labels for each animal in the training images. This information is rarely relevant for ecology, and obtaining bounding box labels is costly; consequently, few camera trap projects have such labels. This makes training object detection models impractical for many camera trap projects, although recent work has demonstrated the effectiveness of object detection when bounding box labels are available \cite{schneider2018deep,beery2018rec}.

\subsection{Transfer learning}
Despite not explicitly being trained to do so, deep neural networks trained on image datasets often exhibit an interesting phenomenon: early layers learn to detect simple patterns like edges \cite{krizhevsky2012imagenet}. Such patterns are not specific to a particular dataset or task, but they are general to different datasets and tasks. Subsequent layers detect more complex and more specific patterns to the dataset the network is trained on. Eventually, there is a transition from general features to dataset-specific features, and from simple to complex patterns within the layers of the network \cite{yosinski2014transferable}.

\textit{Transfer learning} is the application of knowledge gained from learning a task to a similar, but different, task \cite{yosinski2014transferable}. Transfer learning is highly beneficial when we have a limited number of labeled samples to learn a new task (for example, species classification in camera trap images when the new project has few labeled images), but we have a large amount of labeled data for learning a different, relevant task (for example, general-purpose image classification). In this case, a network can first be trained on the large dataset and then \textit{fine-tuned} on the target dataset \cite{yosinski2014transferable,norouzzadeh2018automatically}. Using transfer learning, the general features deep neural networks learn on a large dataset can be reused to learn a smaller dataset more efficiently.

\subsection{Active learning}
In contrast to the supervised learning scenario, in which we first collect a large amount of labeled examples and then train a machine learning model, in an \textit{active learning scenario} we have a large pool of unlabeled data and an oracle (e.g.\ a human) that can label the samples upon request. Active learning iterates between training a machine learning model and asking the oracle for \textit{some} labels, but it tries to minimize the number of such requests. The active learning algorithm must select the samples from the pool for the oracle to label so that the underlying machine learning model can quickly learn the requested task.

Active learning algorithms maintain an underlying machine learning model, such as a neural network, and try to improve that model by selecting training samples. Active learning algorithms typically start training the underlying model on a small, randomly-selected labeled set of data samples. After training the initial model, various criteria can be employed to select the most informative unlabeled samples to be passed to the oracle for labeling \cite{settles2009active}. Among the most popular query selection strategies for active learning are model uncertainty \cite{lewis1994sequential}, query-by-committee (QBC) \cite{seung1992query}, expected model change \cite{settles2008analysis}, expected error reduction \cite{guo2007optimistic}, and density-based methods \cite{settles2008analysis,sener2017active}. For more information on the criteria we use in this paper, refer to the Supplementary Information (SI) sec. \ref{si_active}. After obtaining the new labels from the oracle and retraining the model, the same active learning procedure can be repeated until a pre-determined number of images have been labeled, or until an acceptable accuracy level is reached. Algorithm \ref{activelearning} summarizes an active learning workflow in pseudocode.

\begin{algorithm}
\caption{Active learning procedure}
\label{activelearning}
\begin{algorithmic}[1]
\State Start from a small, randomly-selected labeled subset of data
\While{Stopping criteria not met}
\State Train the underlying model with the available labeled samples
\State Compute a selection criterion for all the samples in the unlabeled pool
\State Select n samples that maximize the criterion
\State Pass the selected samples to the oracle for labeling
\State Gather the labeled samples and add them to the labeled set  
\EndWhile
\end{algorithmic}
\end{algorithm}

\subsection{Embedding learning}
\label{embeddingLearning}
An \textit{embedding function} maps data from a high-dimensional space to a lower-dimensional space, for example from the millions of pixel values in an image (high-dimensional) to a vector of dozens or hundreds of numeric values. Many dimensionality reduction algorithms such as PCA \cite{martinez2001pca} and LDA \cite{martinez2001pca}, or visualization algorithms like t-SNE \cite{maaten2008visualizing}, can be regarded as embedding functions. 

Deep neural networks are frequently used for dimensionality reduction: the input to a deep network often has many values, but layers typically get smaller throughout the network, and the output of a layer can be viewed as a reduced representation of the network's input. In this paper, we use two common methods to train a deep neural network to produce useful embeddings:
\begin{enumerate}
\item We learn an embedding in the course of training another task (e.g., image classification). Here we follow common practice and train a deep neural network for classification with a \emph{cross-entropy loss} and use the activations of the penultimate layer after training as the embedding. Cross-entropy is the most common loss function used for classification problems \cite{goodfellow2016deep}.
\item We learn an embedding that specifically maps samples from the same class to nearby regions in the learned embedding space \cite{koch2015siamese,schroff2015facenet}. Triplet loss \cite{schroff2015facenet} is a popular loss function to accomplish this goal. For more details on triplet loss, refer to SI sec. \ref{si_triplet}. 
\end{enumerate}

We thus have two experimental treatments regarding embedding learning: one with a cross-entropy loss and another with a triplet loss.

\subsection{Datasets}
Three datasets will be used for training and evaluating models in our experiments: Snapshot Serengeti, eMammal Machine Learning, NACTI, and Caltech Camera Traps.

\subsubsection*{Snapshot Serengeti}
The Snapshot Serengeti dataset contains 1.2 million multi-image sequences of camera trap images, totaling 3.2 million images. Sequence-level species, count, and other labels are provided for 48 animal categories by citizen scientists \cite{swanson2015snapshot}. Approximately 75\% of the images are labeled as empty. Wildebeest, zebra, and Thomson's gazelle are the most common species.

\subsubsection*{eMammal Machine Learning}
eMammal is a data management platform for both researchers and citizen scientists working with camera trap images.  We worked with a dataset provided by the eMammal team specifically to support machine learning research, containing over 450,000 images and over 270 species from a diverse set of locations across the world \cite{forrester2013emammal,emammal}.

\subsubsection*{NACTI}
The North America Camera Trap Images (NACTI) dataset \cite{nacti} contains 3.7 million camera trap images from five locations across the United States, with labels for 28 animal categories, primarily at the species level (for example, the most common labels are cattle, boar, and red deer). Approximately 12\% of images are labeled as empty.
\subsubsection*{Caltech Camera Traps}
The Caltech Camera Traps (CCT) dataset \cite{cct} contains 245 thousand images from 140 camera traps in the Southwestern United States. The dataset contains 22 animal categories. The most common species are opossum, raccoon, and coyote. Approximately 70\% of the images are labeled as empty.

\section{Methods}
\label{sec:methods}
In this paper, we propose a pipeline to tackle several of the major roadblocks preventing the application of deep learning techniques to camera trap images.  Our proposed pipeline takes advantage of transfer learning and active learning to concurrently help with the transferability issue, multi-species images, inaccurate counting, and limited-data problems. In this section, we explain the details of our procedure and the motivations for each step.

\subsection{Proposed pipeline}
Our pipeline begins with running a pre-trained object detection model, based on the Faster-RCNN object detection algorithm \cite{ren2015faster}, over the images. The pre-trained model is available to download \cite{megadetectorurl}. We utilized version 2 of the model. This version of the model has only one class -- \textit{animal} -- and was trained on several camera trap datasets that have bounding box annotations available. We threshold the predictions of the model at 90\% confidence and do not consider any detection with less than 90\% confidence. The pre-trained object detection model accomplishes three related tasks:

\begin{enumerate}
\item It can tell us if an image is empty or contains animals; any image with no detections above 90\% confidence is marked as empty.
\item It can count how many animals are in an image; we count animals by summing the number of detections above 90\% confidence.
\item By localizing the animals, it can be employed to crop the images to reduce the amount of background pixels; we crop detections above 90\% confidence and use these cropped images to recognize species in the next steps of the pipeline.
\end{enumerate}

After running the object detection model over a set of images, we have already marked empty images, counted animals in each image, and gathered the crops to be further processed. Image classification models require fixed-sized inputs; since crops are variable in size, we resize all the crops to 256 $\times$ 256 pixels regardless of their original aspect ratio using bilinear interpolation. This set of cropped, resized images -- which now contain animals with very little background -- is the data we process with active learning. 

There are two major challenges for applying active deep learning on a large, high-dimensional dataset: (1) We expect to have relatively few labeled images for our target dataset, typically far too few to train a deep neural network from scratch. Consequently, when training a model for a new dataset, we would like to leverage knowledge derived from related datasets (i.e., other camera trap images); this is a form of \textit{transfer learning} \cite{yosinski2014transferable}. (2) Active learning usually requires cycling through the entire unlabeled dataset to find the next best sample(s) to ask an oracle to label. Processing millions of high-dimensional samples to select active learning queries is impractically slow. One could approximate the next best points by only searching a random subset of the data, but that comes at the cost of inefficiency in the use of the oracle's time (i.e., they will no longer be labeling  the most informative images). 

Our proposed method allows us to evaluate all data points in order to find the most significant examples to ask humans to label, while retaining speed. Before processing the crops from a target dataset, we learn an \textit{embedding model} (a deep neural network) on a large dataset, and use this model to embed the crops from our target dataset into a 256-dimensional feature space. The embedding model turns each image into a 256-dimensional \textit{feature vector}. Using this technique we can both take advantage of transfer learning and significantly speed up the active learning procedure. The speedup occurs because when cycling over all data points we already have a low-dimensional feature vector to process, instead of needing to process each high-dimensional input by running it through a neural network. As discussed in sec. \ref{embeddingLearning}, we experiment with two embeddings produced by the cross-entropy and triplet losses, respectively (discussed more in sec. \ref{sec:embed}).

After obtaining the features for each crop in the lower-dimensional space, we have all the necessary elements to start the \textit{active learning loop} over our data. We employ a simple neural network with one hidden layer consisting of 100 neurons as our \textit{classification model}. We start the active learning process by asking the oracle to label 1,000 randomly-selected images. We then train our classification model using these 1,000 labeled images. At each subsequent step, we select 100 unlabeled images that maximize our image selection criteria (we will discuss different image selection strategies in sec. \ref{sec:strategies}), and ask the oracle to label those 100 images; the classifier model is re-trained after each step.  Another important step in our active learning algorithm is fine-tuning the embedding model periodically, which we do every 20 steps, starting after 2,000 images have been labeled. 

Our pipeline is presented in pseudocode form as Algorithm \ref{pipeline}.

\begin{algorithm}
\caption{Proposed pipeline}
\label{pipeline}
\begin{algorithmic}[1]
\State Run a pre-trained object detection model on the images
\State Run a pre-trained embedding model on the crops produced by the objection detection model
\State Select 1,000 random images and request labels from the human oracle
\State Run the embedding model on the labeled set to produce feature vectors
\State Train the classification model on the labeled feature vectors
\While{Termination condition not reached}
\State Select 100 images using the active learning selection strategy, pass these to the human oracle for labeling
\State Fine-tune the classification model on the entire labeled set of the target dataset
\If{number of examples \% 2,000 == 0}
\State Fine-tune the embedding model on the entire labeled set of the target dataset
\EndIf
\EndWhile
\end{algorithmic}
\end{algorithm}

\section{Experiments and results}
As explained above, our suggested pipeline consists of three steps: (1) running a pre-trained detector model on images, (2) embedding the obtained crops into a lower-dimensional space, and (3) running an active learning procedure. In this section, we report the results of our pipeline and analyze the contribution of these steps to the overall results. For these results, the eMammal Machine Learning dataset is used to train the embedding model, and the target dataset is Snapshot Serengeti. We chose eMammal Machine Learning for training our embedding because it is the most diverse of the available datasets and thus likely provides the most general model for applying to new targets. We chose Snapshot Serengeti as our target dataset to facilitate comparisons with the results presented in \cite{norouzzadeh2018automatically}.

\subsection{Empty vs. animal}
We run a pre-trained object detection model on the target dataset, and we consider images containing any detections above 90\% confidence to be an image containing an animal (i.e. non-empty). The remaining images (containing no detection with more than 90\% confidence) are marked as empty images. As the results in Table \ref{tab:EFCE} show, the detector model has 91.71\% accuracy, 84.47\% precision, and 84.23\% recall. Compare these results with those of \cite{norouzzadeh2018automatically} which are 96.83\% accuracy, 97.50\% precision, and 96.18\% recall. We stress that that this accuracy came``for free", without manually labeling any image for the target dataset, while Norouzzadeh et al. \cite{norouzzadeh2018automatically} used 1.6 million labeled images from the target dataset to obtain their results. The pre-trained model was trained on the few camera trap datasets for which bounding box information exists; we expect this accuracy to improve as the pre-trained object detection model gets trained on larger, more diverse datasets.

\begin{table}[h!]
	\caption{The confusion matrix for the pre-trained object detection model applied to the Snapshot Serengeti dataset}
	\centering
	\setlength{\unitlength}{0.10in}
	\begin{tabular}{p{3cm}p{2cm}p{2cm}p{2cm}}
		\multicolumn{4}{c}{Model Predictions}\\
		\hline
		&&\textbf{Empty}&\textbf{Animal}\\
		\hline
		\multirow{2}{*}{Ground Truth Labels}&\textbf{Empty}&2,219,404&131,288\\
		&\textbf{Animal}&133,769&714,276\\
		\hline
	\end{tabular}
	\label{tab:EFCE}
\end{table}

\subsection{Counting}
Using a pre-trained object detection model allows us to not only distinguish empty images from images containing animals, but also to count the number of animals in each image. This simply means counting the number of bounding boxes with more than 90\% confidence for each image. This straightforward counting scheme can give us the exact number of animals for 72.4\% of images, and the predicted count is either exact or within one bin for 86.8\% of images (following \cite{swanson2015snapshot} we bin counts into 1, 2, ..., 9, 10, 11-50, 51+). Comparing to counting accuracy in Norouzzadeh et al., both the top-1 accuracy and the percent within +/- 1 bin are slightly improved, and this improvement comes ``for free" (i.e., without \emph{any} labeled images from the target dataset).

\subsection{Species identification}
After eliminating empty images and counting the number of animals in each image, the next task is to identify the species in each image. As per above, for species identification, we first embed the cropped boxes into a lower-dimensional space, then we run an active learning algorithm to label the crops. In the next three subsections, we discuss the details of each step and compare several options for implementing them.

\subsubsection{Embedding spaces}
\label{sec:embed}
As described above, we experimented with both (1) using features of the last layer of an image classification network trained on a similar dataset using cross-entropy loss, and (2) using features obtained from training a deep neural network using triplet loss \cite{schroff2015facenet,hermans2017defense} on a similar dataset.
We used the ResNet-50 architecture \cite{he2016deep} for both treatments; only the loss function differs between these methods. After extracting the features with both techniques, we run the same active learning strategy on both sets of features. For these experiments, we chose the active learning strategy that worked the best in our experiments (Sec. \ref{sec:strategies}), which is the ``k-Center'' method \cite{sener2017active} (Sec. \ref{sec:strategies} provides a brief description of the method).

\begin{figure}[h]
	\setlength{\unitlength}{0.5in}
	\centering 
	\subfloat[Softmax cross-entropy]{{\includegraphics[width=0.85\textwidth,clip]{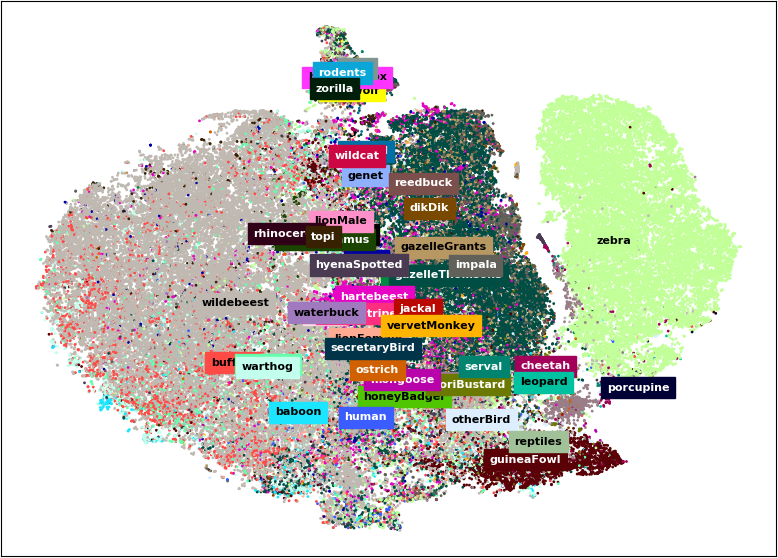} }}%
	\hspace{1pt}
    \subfloat[Triplet]{{\includegraphics[width=0.85\textwidth,clip]{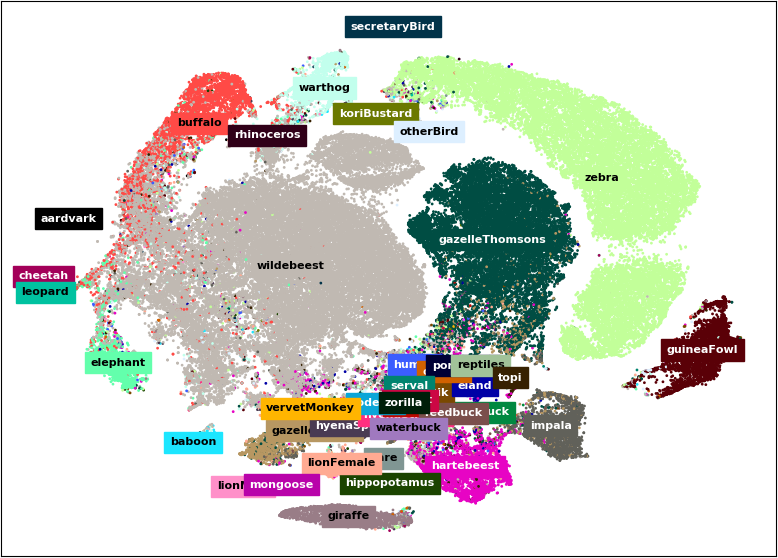} }}%
	\caption{t-SNE visualization of 100,000 randomly selected crops from the Snapshot Serengeti dataset with the embedding spaces produced by the (a) softmax cross-entropy loss and (b) triplet loss. The embedding based on triplet features shows a more intuitive, intelligent, separated distribution of species in the embedding space.
	}
	\label{fig:embed}
\end{figure}

\begin{figure}[h]
	\setlength{\unitlength}{0.5in}
	\centering 
	\includegraphics[width=1\textwidth]{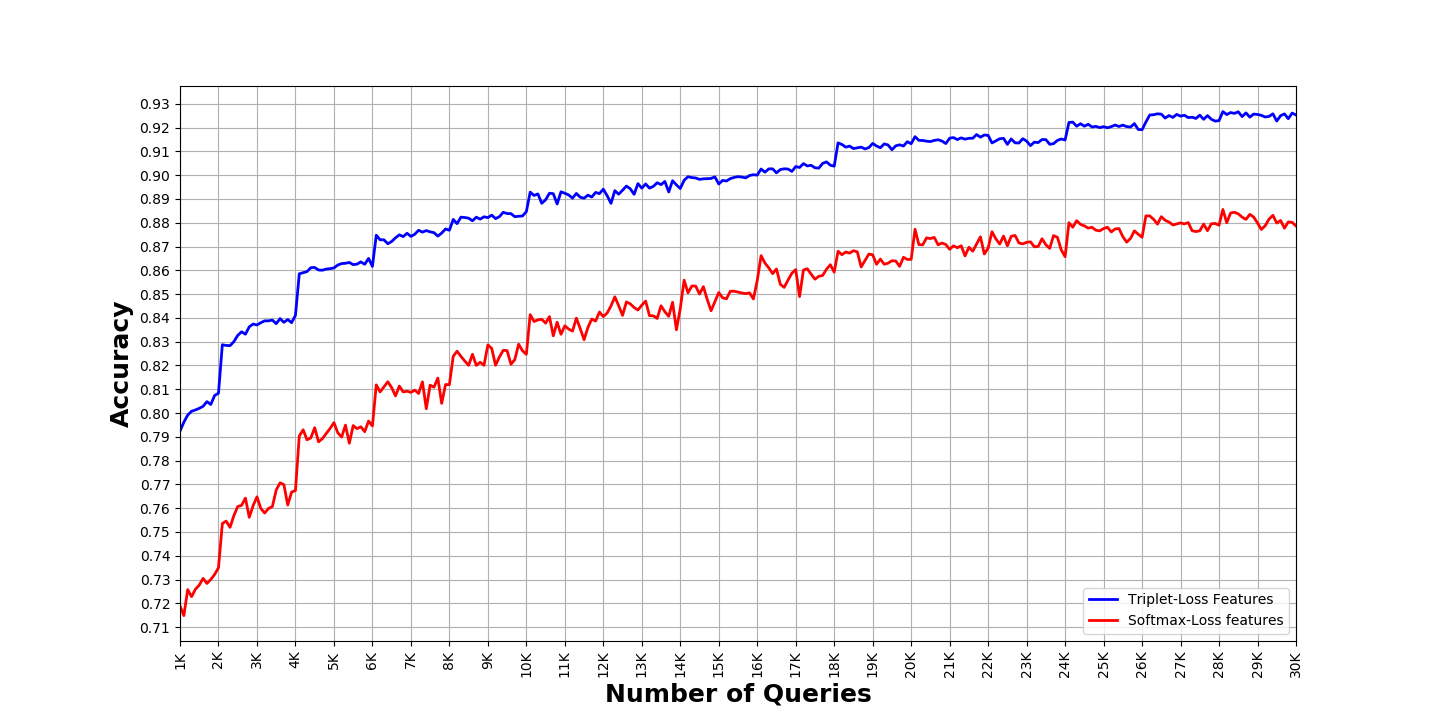}
	\caption{The accuracy of an active learning process using triplet loss features vs. using softmax cross-entropy loss features. Triplet loss features work better, but the gap closes as number of queries increases.
	}
	\label{fig:loss}
\end{figure}

After only 25,000 labels (a low number by deep learning standards), we achieved 85.23\% accuracy for the features extracted from the last layer of a classification model and 91.37\% accuracy with the triplet loss features. Fig. \ref{fig:embed} depicts the t-SNE visualization of the learned embedding space. These results indicate that using triplet loss to build the embedding space provides better accuracy than features derived from an image classification model. As mentioned above, fine-tuning the embedding model periodically by using the obtained labels has a significant positive effect on improving accuracy. The jumps in accuracy (Fig. \ref{fig:loss}, \ref{fig:active}, and \ref{fig:crop}) at 2K, 4K, 6K, ..., 28K clearly depict the advantage of fine-tuning the embedding model periodically. The results suggest it is better to use triplet loss with limited data.

The performance benefits of triplet loss likely stem from additional constraints placed on the embedding.  Cross-entropy loss uses each sample independently, but in triplet loss, we use combinations of labeled samples (i.e., triplets), and we may reuse each sample in many triplets. For example, consider having 1,000 labeled images (10 classes, 100 samples each). In the cross-entropy loss scenario, we have 1,000 constraints over the weights of the network we optimize. In the triplet loss scenario, we can make up to 1,000 (choice of the anchor sample) $\times 99$ (choice of the positive sample) $\times 900$ (choice of the negative sample) $=89,100,000$ constraints over the parameters. Using triplets thus provides 8,910 times more constraints than cross-entropy loss. These additional constraints help find a more informed embedding, and that improvement is qualitatively evident in Fig. \ref{fig:embed}. Of course, not all the possible combinations for triplet loss are useful, because many of them are easily satisfied. That is why we mine for hard triplets during training (Sec. \ref{si_triplet}).
As we fine-tune the embedding model with far more labeled images, we expect the gap between the performance of cross-entropy loss and triplet loss to get smaller, because eventually both methods have sufficient constraints to learn a good embedding.

\subsubsection{Active learning strategies}
\label{sec:strategies}
\begin{figure}[h]
	\setlength{\unitlength}{0.5in}
	\centering 
	\includegraphics[width=1\textwidth]{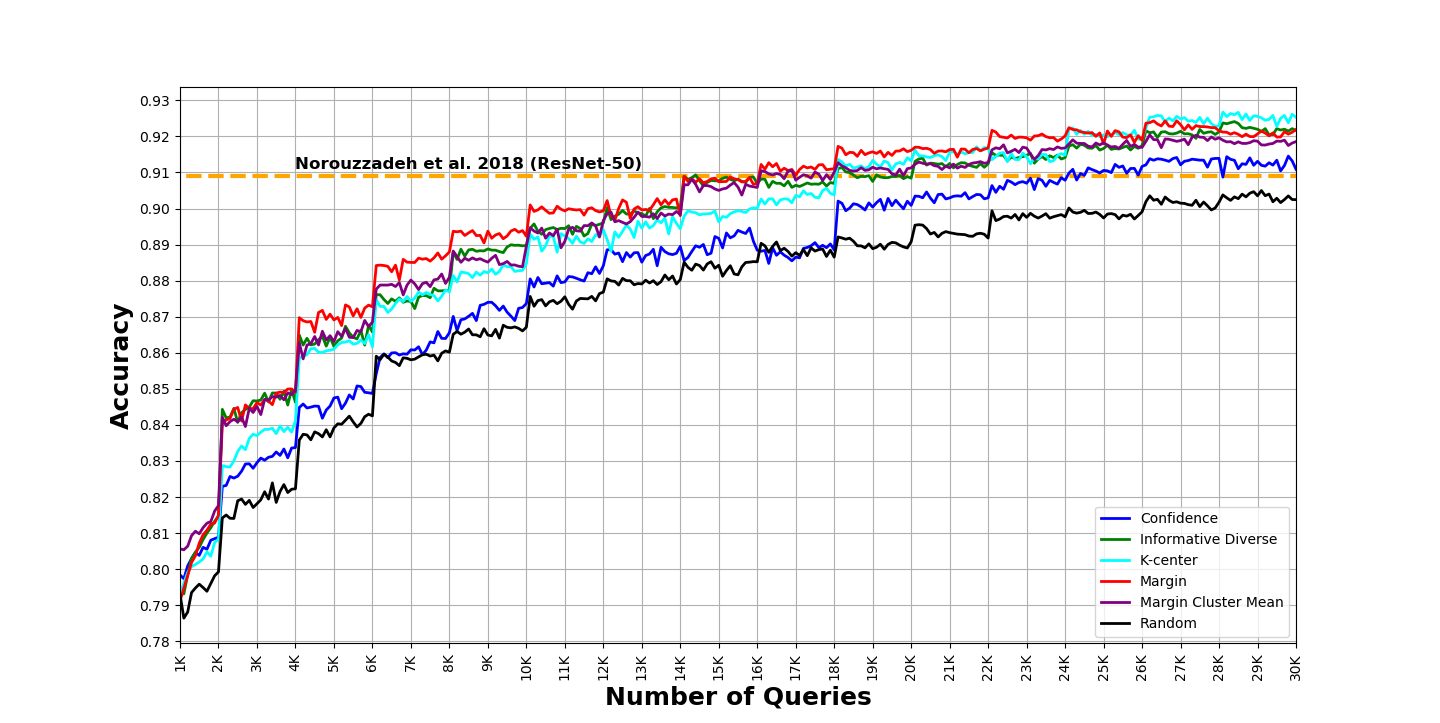}
	\caption{Performance of different active learning query strategies using triplet loss features over the Snapshot Serengeti dataset. k-Center achieves the best accuracy at 30,000 queries.
	}
	\label{fig:active}
\end{figure}

Different strategies can be employed to select samples to be labeled by the oracle. The most naive strategy is selecting queries at random. Here we try five different query selection strategies and compare them against a control of selecting samples at random. In particular, we try model uncertainty criteria (confidence, margin, entropy) \cite{lewis1994sequential}, information diversity \cite{dasgupta2008hierarchical}, margin clustering \cite{xu2003representative}, and k-Center \cite{sener2017active}. For all of these experiments, we use triplet loss features. Considering the expensive computational time and cost of each experiment, we only ran each experiment once. All the active learning strategies show performance improvement over the random baseline (Fig. \ref{fig:active}).  The highest accuracy is achieved with the k-Center strategy, which reaches 92.2\% accuracy with 25,000 labels. The k-Center method selects a subset of unlabeled samples such that the loss value of the selected subset is close to the ``expected'' loss value of the remaining data points \cite{sener2017active}. At 14,000 labels, we match the accuracy of Norouzzadeh et al. for the same architecture; compared to the 3.2 million labeled images they trained with, our results represent over a 99.5\% reduction in labeling effort to achieve the same results.

\subsubsection{Crops vs. full-image classification}
\begin{figure}[h]
	\setlength{\unitlength}{0.5in}
	\centering 
	\includegraphics[width=1\textwidth]{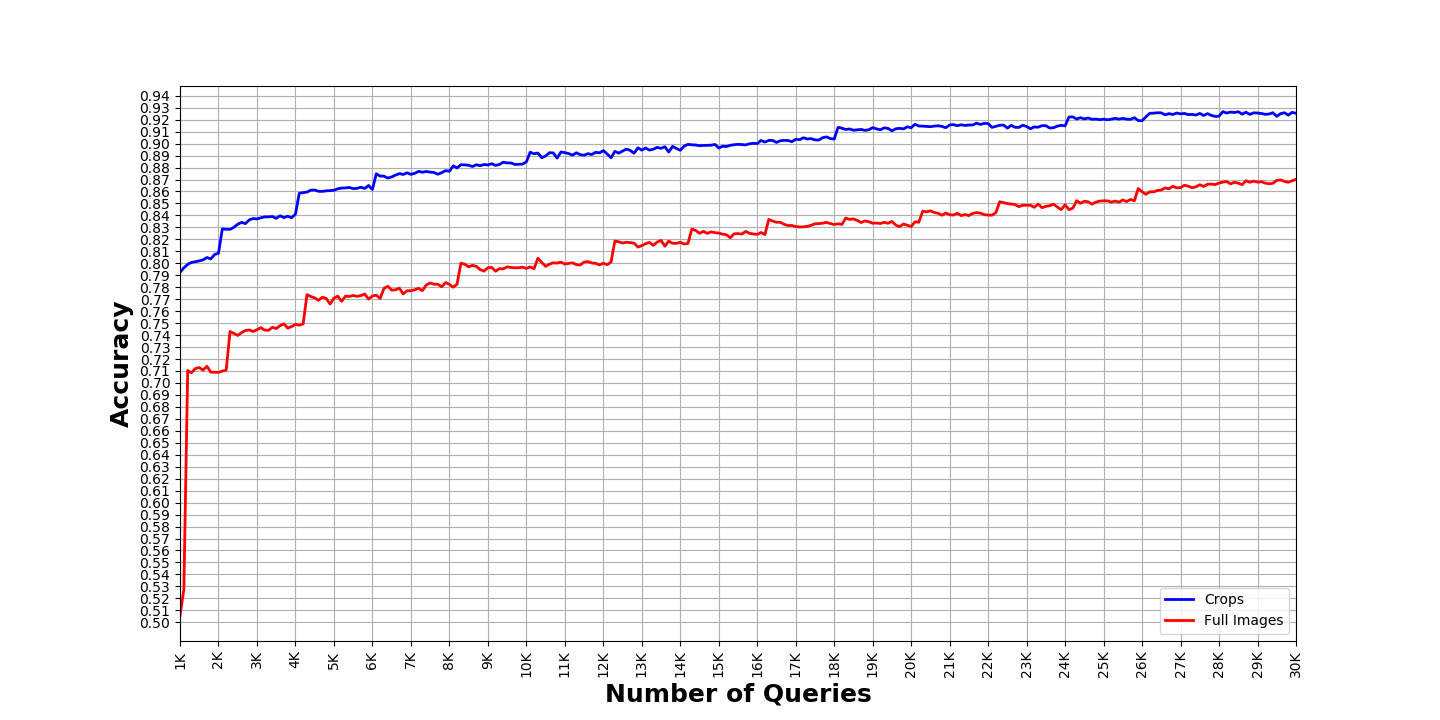}
	\caption{The accuracy of k-Center active learning  using triplet loss features over crops vs. k-Center active learning using triplet loss features over full images on the Snapshot Serengeti dataset. Crops provide a substantial increase in accuracy.
	}
	\label{fig:crop}
\end{figure}

As per above, we identify species in images that have been cropped by the object detection model. To assess the contribution of this choice to our overall accuracy, we also tried to classify species using full images. Fig \ref{fig:crop} shows that using crops produces significantly better results on our data than using full images. This is likely because cropped images eliminate background pixels, allowing the classification model to focus on animal patterns.

\section{Further improvement}
\label{sec:improvement}
This paper demonstrates the potential to significantly reduce human annotation time for camera trap images via active learning.  While we have explored some permutations of our active learning pipeline, we have not extensively explored the space of parameters and algorithmic design choices within this pipeline. We believe there are at least three mechanisms by which our results could be improved.

\begin{enumerate}
\item Every deep learning algorithm has numerous \textit{hyperparameters}, options selected by the data scientist before machine learning begins. For this paper, we used well-known values of hyperparameters to train our models. Tuning hyperparameters is likely to improve results. In particular, we only used the ResNet-50 architecture for embedding and a simple two-layer architecture for classification. Further probing of the architecture space may improve results.
\item We use a pre-trained detector, and we do not modify this model in our experiments. However, if we also obtain bounding box information from the oracle during the labeling procedure, we can fine-tune the detector model in addition to the embedding and classification models.
\item After collecting enough labeled samples for a dataset, it is possible to combine the classification and detection stages into a single multi-class detector model. This may improve accuracy, but almost certainly will improve computational efficiency when applying the model to new datasets.
\end{enumerate}

\section{Conclusion}
\label{sec:conclusion}
Our proposed pipeline may facilitate the deployment of large camera trap arrays by reducing the annotation bottleneck (in our case, by 99.5\%), increasing the efficiency of projects in wildlife biology, zoology, ecology, and animal behavior that utilize camera traps to monitor and manage ecosystems. 

This work suggests the following three conclusions:
\begin{enumerate}
\item Object detection models facilitate the handling of multiple species in images and can effectively eliminate background pixels from subsequent classification tasks. Thus, detectors can generalize better than the image classification models to other datasets.
\item The embeddings produced by a triplet loss outperform those from a cross-entropy loss, at least in case of having limited data.
\item
\textit{Active learning}--machine learning methods that leverage human expertise more efficiently by selecting example(s) for labeling--can dramatically reduce the human effort needed to extract information from camera trap datasets. 
\end{enumerate}

\bibliographystyle{unsrt}

\bibliography{main}

\setcounter{table}{0}
\renewcommand{\thetable}{S\arabic{table}}%
\setcounter{figure}{0}
\renewcommand{\thefigure}{S\arabic{figure}}%
\setcounter{section}{0}
\renewcommand{\thesection}{S\arabic{section}}%
\section*{Supplementary Information}
\section{Triplet loss}
\label{si_triplet}
Triplet loss is originally designed for problems with a variable number of classes such as human face recognition \cite{schroff2015facenet}. Recent studies \cite{hermans2017defense} showed the effectiveness of triplet loss in learning a useful encoding. Triplet loss tries to put samples with the same label nearby in the embedding space, while samples with different labels are mapped to distant points in the embedding space. To train a network using triplet loss, we arrange the labeled examples into triplets. Each triplet consists of a baseline sampled image (the anchor), another sampled image with the same class as the anchor (positive), and a sampled image belonging to a different class (negative). For a distance metric $d$ and a triplet (A, P, N), triplet Loss is defined as:

\begin{equation}
\label{triplet_loss_eq}
L= max(d(A, P) - d(A,N) + margin, 0)
\end{equation}

In Eq. \ref{triplet_loss_eq}, \textit{margin} is a hyperparameter specifying the minimum acceptable difference between $d(A, P)$ and $d(A, N)$. According to the definition of triplet loss, we have three types of triplets: (1) \textit{easy triplets} which already satisfy the condition of triplet loss (i.e., the negative sample is much further than the positive sample to the anchor) and thus have a loss of zero, (2) \textit{semi-hard triplets} in which $d(A,N)>d(A,P)$ but $d(A,N)<d(A,P) + margin$, and (3) \textit{hard triplets} in which the negative sample is closer to the anchor than the positive sample. Easy triplets have a loss of zero and thus have no effect on training the weights of the network. Therefore, we omit them when arranging the triplets. Various strategies could be utilized to form the triplets such as choosing the hardest negative (the negative sample with maximal loss) or randomly choosing a hard or semi-hard negative for each pair of anchor and positive. Just like the original triplet loss paper \cite{schroff2015facenet}, we use the random semi-hard negative strategy in this paper. 
\section{Active learning selection criteria}
\label{si_active}
Many query selection criteria have been proposed in the literature; for our experiments, we employ two criteria based on model uncertainty (confidence-based and margin-based selection \cite{settles2008analysis}) and three criteria based on identifying dense regions in the input space (informative diverse\cite{dasgupta2008hierarchical}, margin cluster mean\cite{xu2003representative}), and k-Center\cite{sener2017active}. In this section, we summarize each of these criteria. For more details on active learning query selection criteria, refer to \cite{settles2008analysis}.

\subsection{Model uncertainty selection}
Both the confidence-based and margin-based techniques belong to the model uncertainty selection category. The main assumption of these approaches is that when the underlying model is uncertain about predicting a sample, that sample could be more informative than the others.  The uncertainty measure is interpreted from the model's output. 

The confidence-based approach chooses the samples for which the model has the lowest confidence in the most probably class; the margin-based approach chooses the samples with the smallest gap between the model's most confident and second-most confident classes.

\subsection{Density-based selection}
The primary assumption of these criteria is that for learning efficiently, we should not only query the labels of uncertain samples, but should also query those samples which are representative of many inputs, i.e. \textit{dense} regions of the underlying input space. This assumption makes density-based methods more resilient to outliers. The informative diverse technique \cite{dasgupta2008hierarchical} first forms a hierarchical clustering of the unlabeled samples and then selects active learning queries so that the distribution of queries matches the distribution of entire data.   The margin cluster mean criterion \cite{xu2003representative} clusters the samples lying within the margin of an SVM classifier trained on the labeled samples, and then selects the samples at cluster centers for human labeling. The k-center method \cite{sener2017active}, which has the best performance in our experiments, chooses a set of samples such that a model trained over the selected subset performs equally well on the remaining samples. The k-center method achieves this goal by defining the problem of active learning as a core-set selection problem \cite{agarwal2005geometric} and then solving it.
\end{document}